\documentclass[journal,twocolumn]{IEEEtran}
\usepackage{graphicx}
\usepackage{amssymb}
\usepackage{color}
\usepackage{booktabs}

\usepackage{cite}

\ifCLASSINFOpdf
\else
\fi
\usepackage{amsmath}
\begin{document}
%

\title{Large age-gap face verification \\ by feature injection in deep networks}

%
%
%

\author{Simone~Bianco
\thanks{S. Bianco is with the Dipartment of Informatics, Systems and Communication (DISCo), University of Milano-Bicocca, 20126 Milano, Italy e-mail: simone.bianco@disco.unimib.it.}
}

\maketitle

\begin{abstract}
This paper introduces a new method for face verification 
across large age gaps and also a dataset containing variations of age in
the wild, the Large Age-Gap (LAG) dataset, with images ranging from child/young to adult/old. The proposed method exploits a deep convolutional neural network (DCNN) pre-trained for the face recognition task on a large dataset and then fine-tuned for the large age-gap face verification task. Fine-tuning is performed in a Siamese architecture using a contrastive loss function. A feature injection layer is introduced to boost verification accuracy, showing the ability of the DCNN to learn a similarity metric leveraging external features. Experimental results on the LAG dataset show that our method is able to outperform the face verification solutions in the state of the art considered.
\end{abstract}


%
\IEEEpeerreviewmaketitle

\section{Introduction}
Face verification is an important topic in both computer vision, imaging and multimedia. 
Verification accuracy mainly depends on four elements: face pose, facial expression, illumination, and aging \cite{jain2012face}. 
The greatest part of the works in the state of the art studied the face verification problem in constrained scenarios, controlling and fixing one or more of these four elements. 

Recently many researchers achieved or even surpassed human-level performance \cite{chen2013blessing,taigman2014deepface} on face verification benchmark taken in unconstrained environments such as the Labeled Faces in the Wild dataset (LFW) \cite{huang2007labeled}. These results have been made possible thanks to the improvement in
facial landmark detection 
and to the increase of the computational power available to train deep models.
However, the LFW dataset fixes the aging element: it contains large variations in pose, facial expression, and illumination, but contains very little variation in aging.
As people grow, face appearance can be very different, which makes it difficult to recognize people across age. 
The problem is even harder when large age gaps are considered.

To address this problem, in this work a new approach is proposed. Differently from other approaches in the state of the art, the proposed method does not rely on parametric models nor tries to model age progression. The idea is to use deep learning to jointly learn face features that matching faces share, and a similarity metric on top of these features. This is done coupling two deep convolutional neural networks (DCNN) with shared parameters in a Siamese network \cite{bromley1993signature,chopra2005learning} ended with a contrastive loss function. The discriminative power of the network is further improved including a feature injection layer, which fuses externally computed features with the activations of the deepest layers of the DCNN.

{\color{black}
The idea of deep feature fusion has been mainly explored in the video categorization task. 
%
%
One of the earliest work is from Simonyan and Zisserman \cite{simonyan2014two} where they proposed a two-stream ConvNet architecture which incorporated a spatial and a temporal network.
%
Perhaps the most similar work is \cite{jiang2015exploiting} where multi-modal video features are combined (e.g. frame-based features computed by a convolutional neural network, trajectory-based motion descriptors and audio descriptors).
Wang et al. \cite{wang2015action} integrate the advantages of hand-crafted and
deep-learned features: they utilize deep architectures to learn multi-scale convolutional feature maps, and introduce the strategies of trajectory-constrained sampling and pooling to encode deep features into effective descriptors.
Zha et al.\cite{zha2015exploiting} propose a late fusion approach between CNN features (taken at different layers) and Fisher Vectors \cite{perronnin2010improving}. The features are fused using an external classifier and thus not in an end-to-end training, excluding the possibility of backward feedbacks on feature extraction.
Ng et al. \cite{ng2015beyond} investigated the combination of Long Short Term Memory
(LSTM) networks \cite{gers2003learning} with optical flow.
%
Park et al.\cite{parkcombining} propose a multiplicative fusion method for combining multiple CNNs trained on different sources. 
}

The contributions of this work are summarized as follows:
\begin{itemize}
\item[-] A new large-scale Large Age-Gap (LAG) dataset is collected, that includes images in the wild of 1,010 international celebrities spanning large age gaps.
\item[-] A new DCNN architecture is proposed, including a feature injection layer that fuses external features with the activations of the deepest DCNN layers.
\item[-] Extensive experiments are conducted on LAG and show that the proposed DCNN architecture can outperform state-of-the-art methods.
\end{itemize}

The remaining sections are organized as follows: Section \ref{sec:related} reviews the related works on face recognition, age-invariant face recognition and existing face datasets. Section \ref{sec:proposed} describes the proposed method, while Section \ref{sec:lag} introduces the Large Age-Gap (LAG) dataset. Experiments are presented in Section \ref{sec:experiments}.
Finally, Section \ref{sec:conclusion} draws the conclusions and discusses future works.

\section{Related works}
\label{sec:related}
\subsection{Face Recognition}
Face recognition has been investigated for a
long time in many different works. 
One of the earliest works is that of Turk and Pentland where they 
introduced the idea of eigenface \cite{turk1991face}. 
Ahonen et al. \cite{ahonen2006face} explored the use of a texture descriptor, i.e. local binary
pattern (LBP), for the face recognition task. 
Wright et al. \cite{wright2009robust} cast face recognition problem as one of classifying among multiple linear regression models via sparse signal representation, showing a high degree of robustness against face occlusions.
%
Chen et al. \cite{chen2013blessing} proposed a high dimensional version of LBP (HDLBP) and studied the performance of face feature as a function of dimensionality, showing that high dimensionality is critical to achieve high performance. 
%

Recently there have been many works exploiting deep learning for face recognition.
Results obtained by Taigman et al. \cite{taigman2014deepface} and by Sun et al. \cite{sun2014deep10k,sun2014deep} using deep convolutional neural networks (DCNNs) reach or even surpass human-level performance on the widely used labeled face in the wild dataset (LFW) \cite{huang2007labeled}.
Although these methods achieve very high performance on face recognition, they do not work well when in presence of age variation, since this information is not used. 

\subsection{Age-Invariant Face Recognition}
The largest part of existing works related to age in face image analysis focus on age estimation and simulation. Only recently researchers have started to work on cross age face recognition. 
Existing works can be grouped into generative and discriminative methods.
Among the first group, some of the approaches build 2D \cite{geng2007automatic} or 3D \cite{park2010age} aging models. These methods rely on parametric models and accurate age annotation or estimation, and thus do not work well in unconstrained scenarios.
Wu et al.\cite{wu2012age} propose a relative craniofacial growth model to model the facial shape change, which is based on the science of craniofacial anthropometry. 
Their method needs age information to predict the new shapes, limiting its applicability since this information is not always available. 

Among the works based on a discriminative approach Li et al. \cite{li2011discriminative} use multi-feature discriminant analysis (MFDA) to process in a unified framework the two local feature spaces generated by the two different local descriptors used, i.e. SIFT and multi-scale LBP. 
Gong et al. \cite{gong2013hidden} proposed a method separating the HOG local feature descriptor 
into two latent factors using hidden factor analysis: an identity factor that is age-invariant and an age factor affected by the aging process.
Chen et al. \cite{chen2014cross,chen2015face} use a data-driven method that leveraging a large-scale image dataset freely available on the internet as a reference set, encodes the low-level feature of a face image with an age-invariant reference space.
Liu et al. \cite{liu2016deep} propose a generative-discriminative approach based on two modules: the aging pattern synthesis module and the aging face verification module. In the aging
pattern synthesis module, an aging-aware denoising auto-encoder is used to synthesize the faces of all the four age groups considered. In the aging face verification module,
parallel CNNs are trained based on the synthesized faces and the original faces to predict the verification score.

\subsection{Face datasets}
{\color{black}
Existing face datasets can be divided into two main groups: the former consists of datasets acquired in controlled environments, the latter datasets in unconstrained environments. 
Most of the older datasets belong to the first group, such as FERET [30], Yale,
and CMU PIE. 
The most popular dataset in uncontrolled environment is the LFW \cite{huang2007labeled}, with a total of 13,233 images of 5,749 people extracted from news programs. 
Pubfig \cite{kumar2009attribute} has been collected with the aim of providing a larger number of images for each individual, and it contains 58,797 images
of 200 identities. The largest dataset available is the CasiaWebFace dataset \cite{yi2014learning} with a total of 986,912 images of 10,575 people.
All the above datasets can be used only for face recognition and verification tasks, since there is almost no age variation.
Concerning age estimation and face recognition
across age, the most used datasets are FGNet \cite{cootes2008fg} and MORPH \cite{ricanek2006morph}. The former is composed of a total of 1,002 images of 82
people with age range from 0 to 69 and an age gap up to 45 years. The latter contains 55,134
images of 13,618 people with age range from 16 to 77 and an age gap up to 5 years.
%
Recently the CACD dataset has been collected \cite{chen2014cross,chen2015face} crawling the web suing as query 2,000 celebrities names for a total of 163,446 images. For a subset of 200 identities images are manually checked and the noisy ones have been removed. Age ranges from 14 to 62 and age gap is up to 10 years.
Very recently the CAFE dataset has been collected \cite{liu2016deep}. It is the first permitting a study on face verification with large age gaps. It is composed of 4,659 images of 901 people and, due to the way images are collected it does not contain precise age information.
A summary of the comparison between existing datasets is reported in Table \ref{tab:LAGstatistics}.
} 

\section{The proposed method}
\label{sec:proposed}
\begin{figure*}
\centering
\includegraphics[width=2.00\columnwidth]{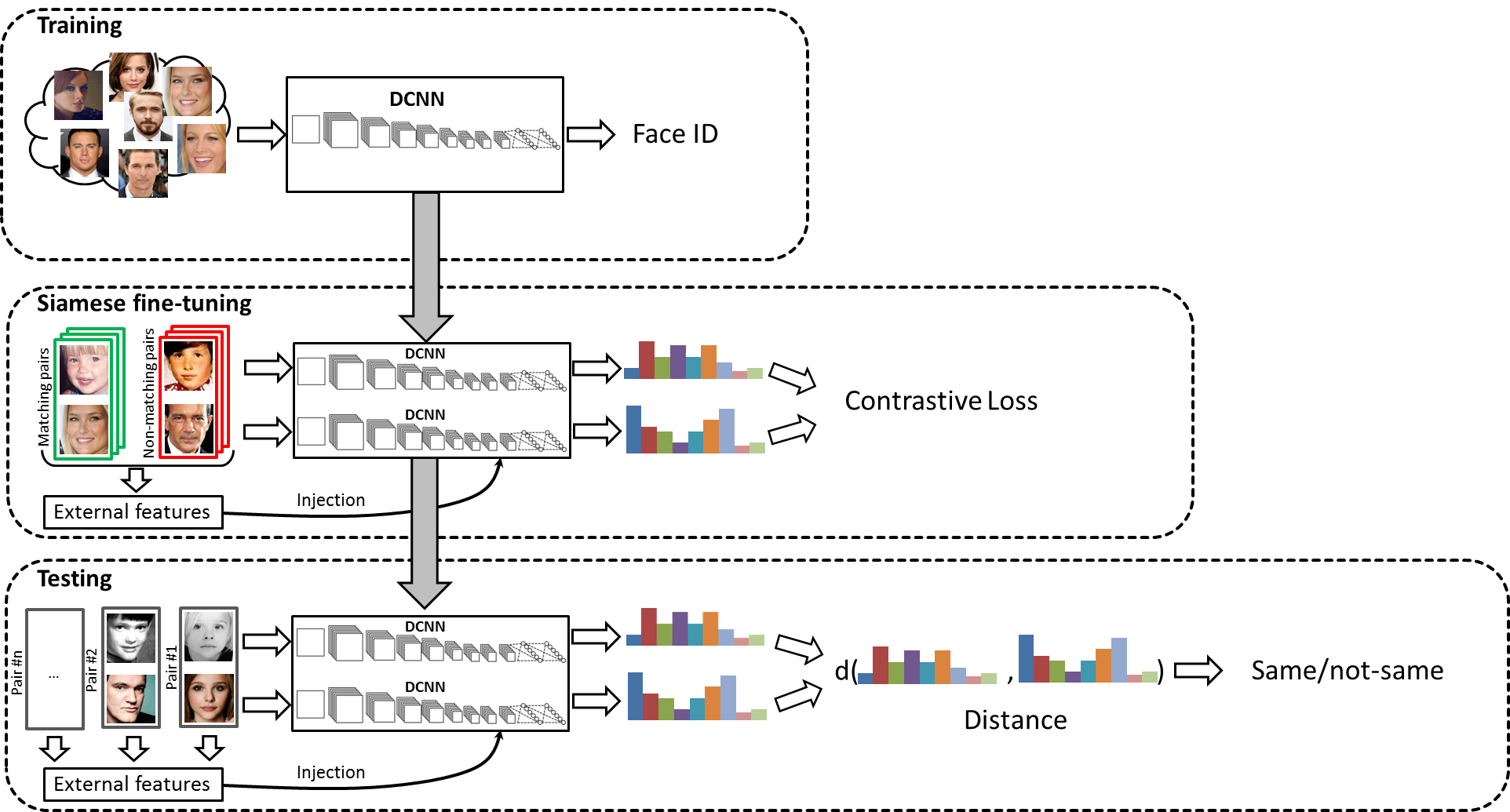}%
\caption{Overview of the proposed method for large age-gap face verification.}%
\label{fig:completePipe}%
\end{figure*}

Figure \ref{fig:completePipe} gives an overview of the proposed method. 
First face and landmark detection are performed on CASIA-WebFace and Large Age-Gap (LAG) database to localize and align each face to a reference position. 
Next, a DCNN is trained on the CASIAWebFace \cite{yi2014learning} for the face recognition task. 
The DCNN is then fine-tuned on the LAG using a contrastive loss in a Siamese architecture in which pre-computed external features are injected in the fully connected layer. 


\subsection{Image preprocessing}

For each image in the database, we first apply the widely used Viola-Jones face detector \cite{viola2004robust} to find the face region in the image. 
For each face, we then locate 68 different facial landmarks using a face alignment algorithm \cite{kazemi2014one}. 
After landmark detection, we use the eye locations to align the face images. 
Images are first rotated so that the eyes are horizontally aligned, then scaled so that the distance between eyes is fixed, and finally cropped to a common size of 200 $\times$ 200 pixels.

\subsection{Deep face feature representation}
\label{sec:DCNN}
A DCNN is trained to learn a discriminative representation of faces. 
The chosen architecture is the AlexNet, but others could be used. 
The DCNN is trained on the face recognition task on the CASIAWebFace. 
The dimensionality of the input layer is 200 $\times$ 200 $\times $1 gray-scale images. 
The network includes 5 convolutional layers, 3 pooling layers and 3 fully connected layers. 
Each convolutional layer is followed by a rectified linear unit (ReLU). 
Two local normalization layers are added after each of the first two convolutional layers to mitigate the effect of illumination variations. 
Dropout is used to regularize all the fully connected layers due to the large number of parameters ($fc6=4096$, $fc7=4097$, and $fc8=10575$ equal to the number of different identities in CASIAWebFace). 
The features extracted from the second to last fully connected layer, i.e. $fc7$, are used for face representation after an $L_2$-normalization step. 

\subsection{Feature injection}
\begin{figure*}[tb]%
\centering
\includegraphics[width=0.8\textwidth]{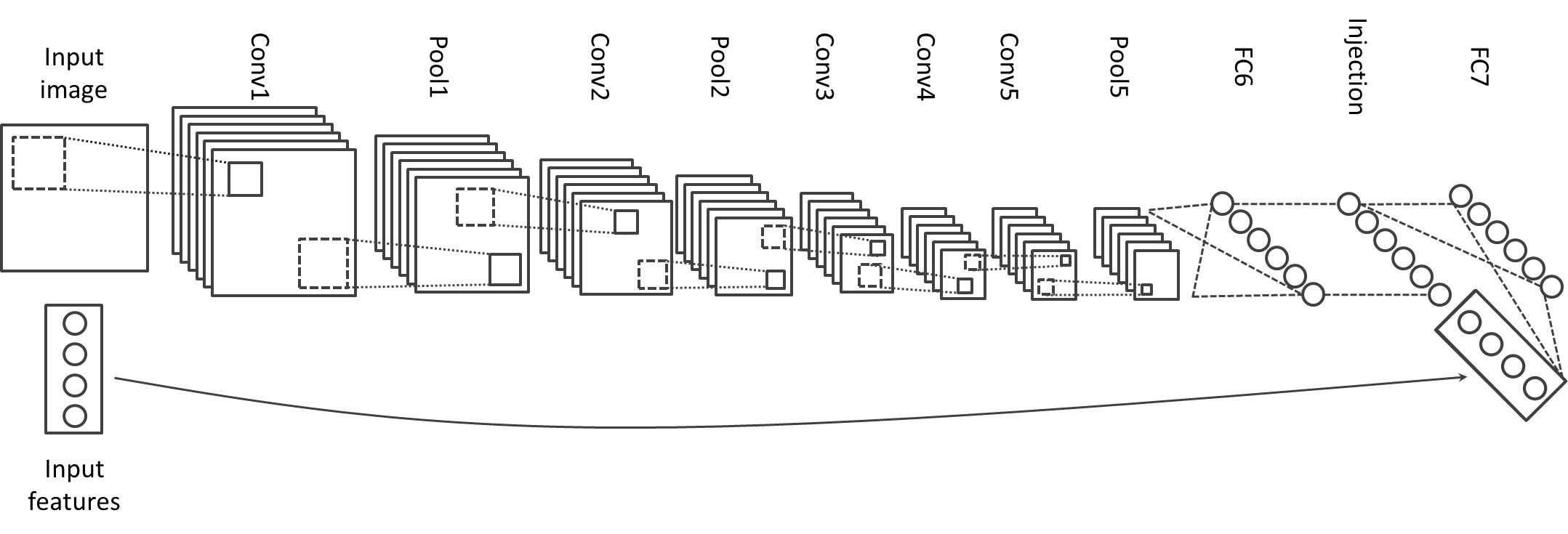}%
\caption{Graphical representation of feature injection in the fully connected layer.}%
\label{fig:injection}%
\end{figure*}

The $L_2$-normalized $fc7$ features are given as input to a set of $n$ face verification methods in the state of the art. Each of them provides as output a distance or confidence score that a pair of images belong to the same identity or not $d_i,i=1,\ldots,n$, 
which are stacked to form the vector $\mathbf{d}=\{d_i\}_{i=1}^n$.
A Siamese DCNN \cite{bromley1993signature,chopra2005learning} is then fine-tuned on the LAG database starting from the net in Section \ref{sec:DCNN} using a contrastive loss function. 
In addition to the DCNN features, a feature injection layer is added to fuse externally computed features with the activations of the deepest layers of the DCNN. In more details, the features $\mathbf{d}$ are injected in the first fully connected layer. i.e. $fc6$. The feature injection is performed in the form of concatenation of the external features $\mathbf{d}$ with the $fc6$'s activations, as represented in Figure \ref{fig:injection} for one side of the Siamese DCNN.
The idea is that the DCNN jointly learns face features that matching faces share, and a similarity metric on top of these features also leveraging external features.

\section{Large Age-Gap dataset (LAG)}
\label{sec:lag}
In order to be able to collect images from a large number of people, the Large Age-Gap dataset (LAG) is constructed with photos of celebrities.
 
Google Image Search is used to collect images by specifying to collect face images. 
In order to collect celebrities images across different ages, we augment the celebrity name with adjectives such as "childhood","adult","now" as keywords. Searches with celebrity names followed by the string "then and now" are also used. YouTube videos of collections of "then and now" celebrities have also been collected.

\begin{figure*}
\centering
\resizebox{\textwidth}{!}{
\begin{tabular}{ccccccccccc}
\includegraphics{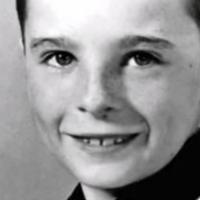} &
\includegraphics{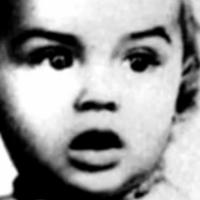} &
\includegraphics{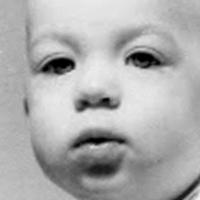} &
\includegraphics{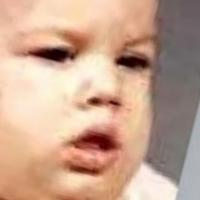} &
\includegraphics{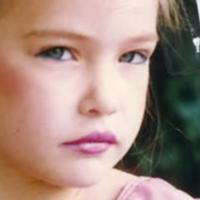} &
\includegraphics{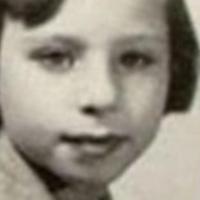} &
\includegraphics{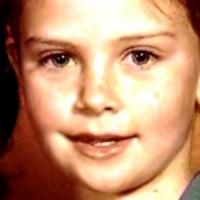} &
\includegraphics{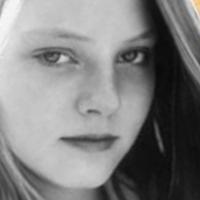} &
\includegraphics{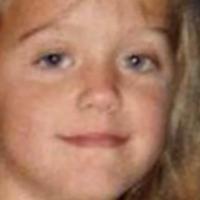} &
\includegraphics{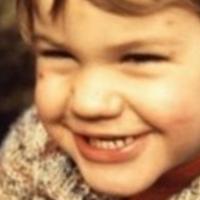} &
\includegraphics{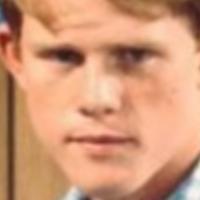} 
\\
\\
\includegraphics{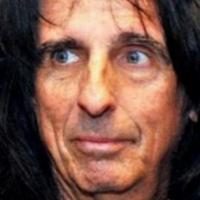} &
\includegraphics{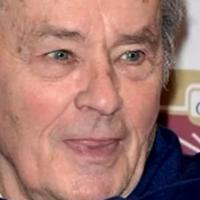} &
\includegraphics{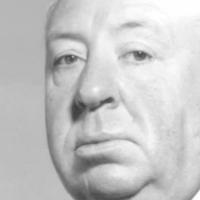} &
\includegraphics{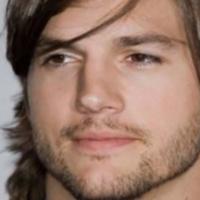} &
\includegraphics{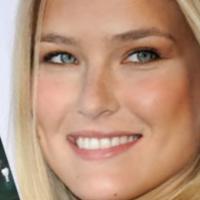} &
\includegraphics{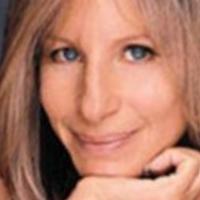} &
\includegraphics{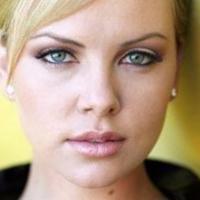} &
\includegraphics{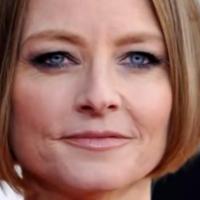} &
\includegraphics{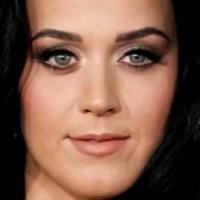} &
\includegraphics{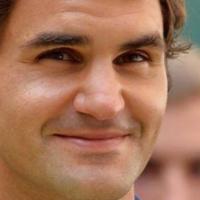} &
\includegraphics{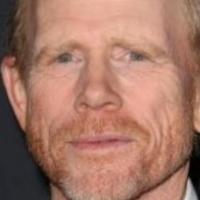} \\
\end{tabular}}
\caption{Examples of face crops for matching pairs in the Large Age-Gap (LAG) dataset.}
\label{fig:LAGexamples}
\end{figure*}

After removing duplicate images, we manually check the images and remove the noisy images in the dataset.
The dataset contains 3,828 images of 1,010 celebrities after removing the noisy images.
For each identity at least one child/young image and one adult/old image are present. 
Starting from the collected images, a total of 5,051 matching pairs has been generated. The same number of non matching pairs has been randomly generated.
Table \ref{tab:LAGstatistics} shows the statistics of the dataset and comparison to other existing face datasets. Some examples of the face crops of the collected images are reported in Figure \ref{fig:LAGexamples}.


\begin{table*}
\caption{The comparison between existing datasets for face verification/recognition.}
\centering
\begin{tabular}{lcrrrcll}
\toprule
Dataset name & Year & \# of images & \# of people & \# images/person & Age info. & Age gap & Publicly available\\
\midrule
LFW   \cite{huang2007labeled} & 2007 & 13,233 &  5,749 &   2.3 & No  & - & Yes\\ 
Pubfig \cite{kumar2009attribute} & 2009 & 58,797 &    200 & 293.9 & No  & - & Yes\\
Casia \cite{yi2014learning} & 2014 & 986,912 & 10,575 &  93.3 & No  & -  & Yes  \\
FGNet \cite{cootes2008fg} & 2008 &  1,002 &     82 &  12.2 & Yes & 0-45 & Yes\\
MORPH \cite{ricanek2006morph} & 2006 & 55,134 & 13,618 &   4.1 & Yes & 0-5 & Yes\\
CACD  \cite{chen2014cross,chen2015face} & 2015 & 163,446 &  2,000 &  81.7 & Yes & 0-10 & Yes\\
CAFE \cite{liu2016deep} & 2016 &4,659 &  901 & 5.2 & No & Large & Not yet\\
Ours (LAG) & this paper &  3,828 &  1,010 &   3.8 & No  & Large  & Yes$^*$ \\
\bottomrule
$*$ after acceptance
\end{tabular}
\label{tab:LAGstatistics}
\end{table*}

\section{Experimental results}
\label{sec:experiments}

In this section we evaluate the performance of large age-gap face verification.
We compare our method with current state-of-the-art features and classifiers for general face verification, such as high dimensional local binary feature (HDLBP) \cite{chen2013blessing}, 
Other similarity metric learning approaches, such as sub-SML \cite{cao2013similarity}, One Shot Similarity Kernel \cite{wolf2009one}, Cosine Similarity \cite{nguyen2011cosine} and Joint Bayesian \cite{chen2012bayesian}. All the similarity learning methods have been trained using the $L_2-$normalized $fc7$ features (extracted with the DCNN trained for the face recognition task on CASIAWebFace). 
We also compare our method with the Cross Age Reference Coding (CARC) \cite{chen2014cross,chen2015face} method, which is designed specifically for cross-age face verification. Since LAG dataset does not provide exact age information, the version without the temporal constraint is used here, i.e. CARC-NT.

To understand the contribution of the feature injection, a Siamese DCNN is fine-tuned on the LAG dataset excluding the injection layer. This also permits a direct comparison of the Siamese DCNN with the other similarity metric learning approaches considered.
Besides feature injection, to have a benchmark, external features are also fused stacking \cite{wolpert1992stacked} linear SVM classifiers as in \cite{hassner2015effective}.

\begin{figure}
\centering
\resizebox{0.95\columnwidth}{!}{
\includegraphics{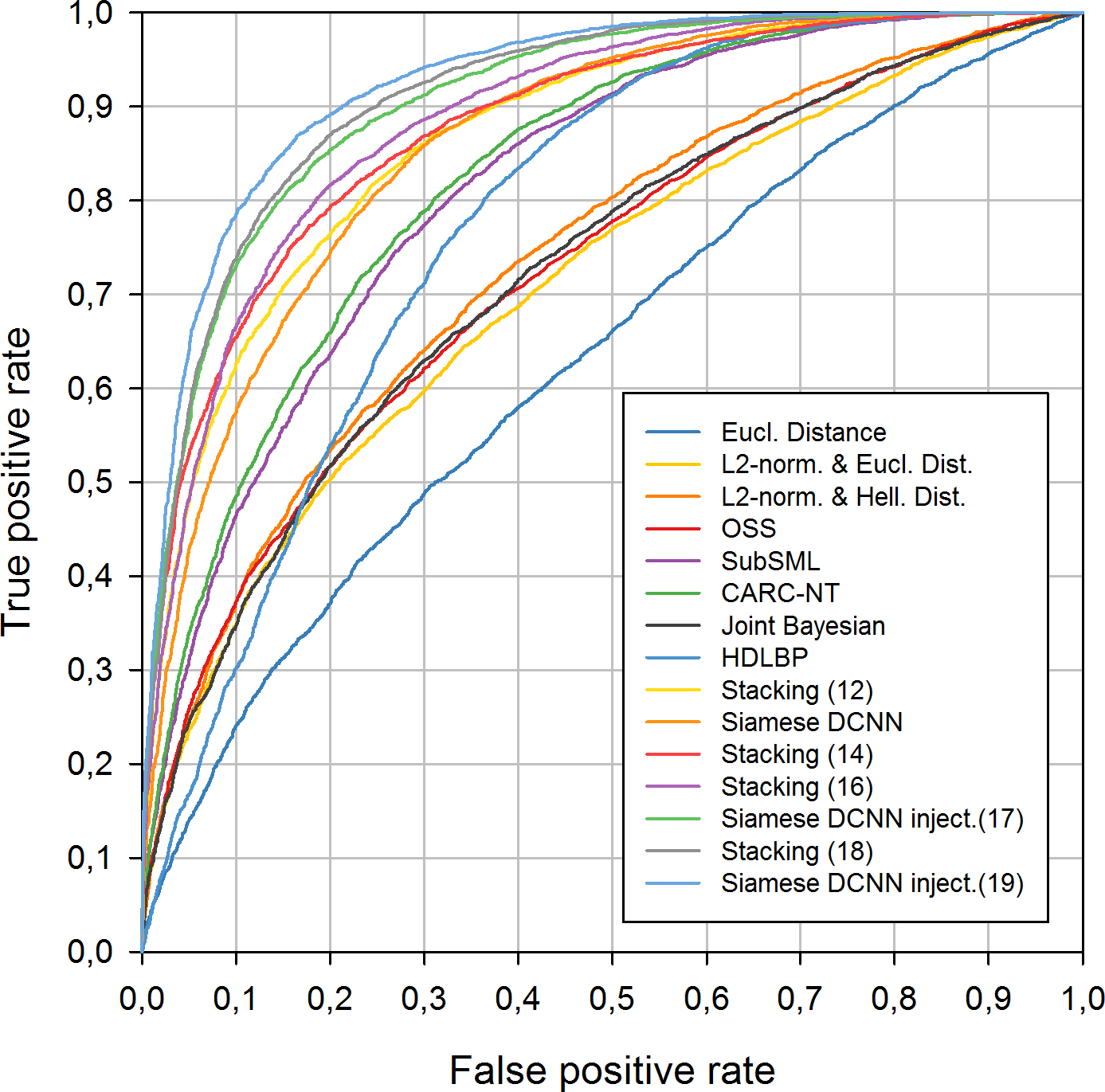}
}
\label{fig:rocLAG}
\caption{ROC curve of the comparison of our method and other baselines on the LAG dataset.}
\end{figure}

Experiments are made using a two-fold cross validation: the LAG dataset has been alphabetically sorted and subjects have been assigned alternately to the first or to the second fold. For each fold, the training set is augmented by considering the four combinations of horizontal flips and original images, as well as tiny amounts of jittering.
Performance are reported as the verification accuracy in Table \ref{tab:resultsLAG2}, and plotted as ROC curves in Figure \ref{fig:rocLAG}.
    
From the results reported in Table \ref{tab:resultsLAG2} it is possible to see that the use of the Siamese DCNN fine-tuned on the LAG dataset to learn a similarity metric is able to outperform all the single methods in the state of the art, outperforming the best one by almost 2.5\%. Enabling the feature injection in the fine-tuning step permits an improvement over the best method in the state of the art of more than 10\%. Feature combination stacking linear SVM classifiers performs 2.3\% worse than our method. The experimental results show the effectiveness of both the use of a Siamese DCNN for similarity metric learning and the usefulness of the feature injection.
An indirect comparison with the method by Liu et al. \cite{liu2016deep} can be done by using HDLBP\cite{chen2013blessing} as a reference: on their CAFE dataset they reported an improvement over HDLBP of 2.5\%. On our LAG dataset our methods outperforms HDLBP by 13.4\%.

In the following we examine more in details the results of the best performing Siamese DCNN with feature injection reported in Table \ref{tab:resultsLAG2} (i.e. ID. 19): its confusion matrix is reported in Table \ref{tab:errorDetails}. 
From the confusion matrix we can evince that our solution is almost equally able to identify non-matching pairs and matching pairs. 

\begin{figure*}
\centering
\resizebox{\textwidth}{!}{
\setlength{\tabcolsep}{2pt}
\begin{tabular}{cccccccccc}
\includegraphics{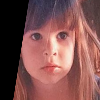} &
\includegraphics{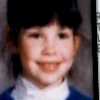} &
\includegraphics{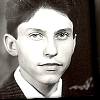} &
\includegraphics{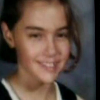} &
\includegraphics{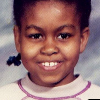} &
\includegraphics{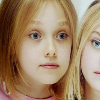} &
\includegraphics{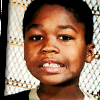} &
\includegraphics{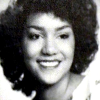} &
\includegraphics{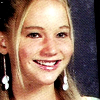} &
\includegraphics{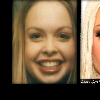} \\
\includegraphics{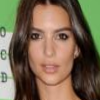} &
\includegraphics{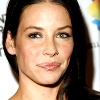} &
\includegraphics{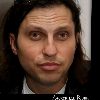} &
\includegraphics{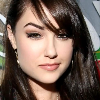} &
\includegraphics{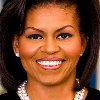} &
\includegraphics{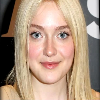} &
\includegraphics{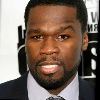} &
\includegraphics{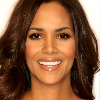} &
\includegraphics{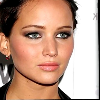} &
\includegraphics{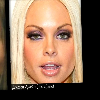} \\
\end{tabular}}
\caption{Examples of correctly classified matching pairs on which our approach is very confident.}
\label{fig:TPconvinced}
\end{figure*}

\begin{figure*}
\centering
\resizebox{\textwidth}{!}{
\setlength{\tabcolsep}{2pt}
\begin{tabular}{cccccccccc}
\includegraphics{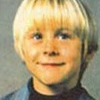} &
\includegraphics{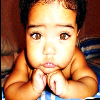} &
\includegraphics{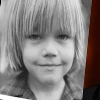} &
\includegraphics{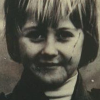} &
\includegraphics{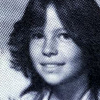} &
\includegraphics{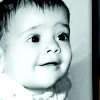} &
\includegraphics{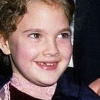} &
\includegraphics{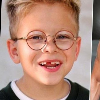} &
\includegraphics{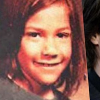} &
\includegraphics{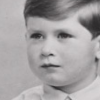} \\
\includegraphics{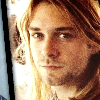} &
\includegraphics{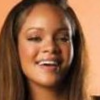} &
\includegraphics{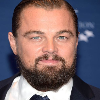} &
\includegraphics{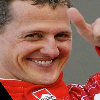} &
\includegraphics{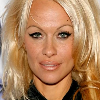} &
\includegraphics{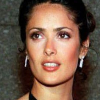} &
\includegraphics{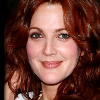} &
\includegraphics{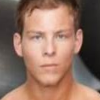} &
\includegraphics{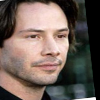} &
\includegraphics{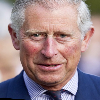} \\
\end{tabular}}
\caption{Examples of correctly classified matching pairs on which our approach is not very confident.}
\label{fig:TPbarely}
\end{figure*}

\begin{figure*}
\centering
\resizebox{\textwidth}{!}{
\setlength{\tabcolsep}{2pt}
\begin{tabular}{cccccccccc}
\includegraphics{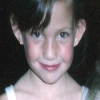} &
\includegraphics{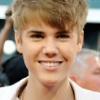} &
\includegraphics{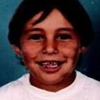} &
\includegraphics{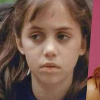} &
\includegraphics{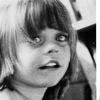} &
\includegraphics{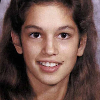} &
\includegraphics{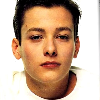} &
\includegraphics{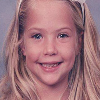} &
\includegraphics{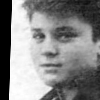} &
\includegraphics{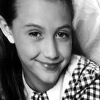} \\
\includegraphics{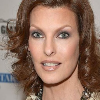} &
\includegraphics{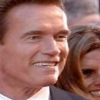} &
\includegraphics{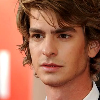} &
\includegraphics{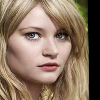} &
\includegraphics{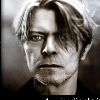} &
\includegraphics{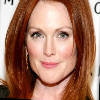} &
\includegraphics{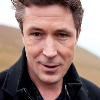} &
\includegraphics{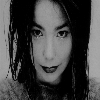} &
\includegraphics{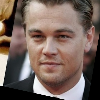} &
\includegraphics{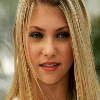} \\
\end{tabular}}
\caption{Examples of errors of our approach: non-matching pairs classified as matching ones (i.e. false positives).}
\label{fig:FP}
\end{figure*}

\begin{figure*}
\centering
\resizebox{\textwidth}{!}{
\setlength{\tabcolsep}{2pt}
\begin{tabular}{cccccccccc}
\includegraphics{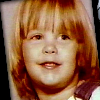} &
\includegraphics{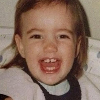} &
\includegraphics{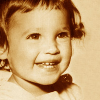} &
\includegraphics{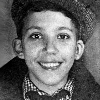} &
\includegraphics{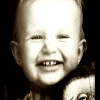} &
\includegraphics{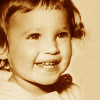} &
\includegraphics{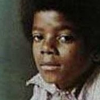} &
\includegraphics{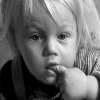} &
\includegraphics{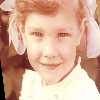} &
\includegraphics{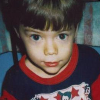} \\
\includegraphics{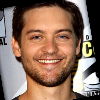} &
\includegraphics{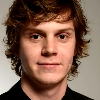} &
\includegraphics{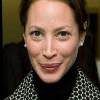} &
\includegraphics{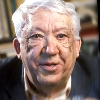} &
\includegraphics{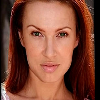} &
\includegraphics{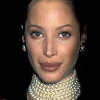} &
\includegraphics{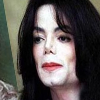} &
\includegraphics{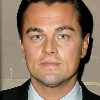} &
\includegraphics{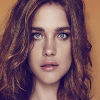} & 
\includegraphics{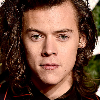}\\
\end{tabular}}
\caption{Examples of errors of our approach: matching pairs classified as non-matching ones (i.e. false negatives).}
\label{fig:FN}
\end{figure*}

In Figure \ref{fig:TPconvinced} some examples of correctly identified matches among those on which our solution is very confident are reported. Some examples of correctly identified matches on which our solution is not so confident are reported in Figure \ref{fig:TPbarely}. 
Some examples of false positives, i.e. non-matching pairs classified as matching ones are reported in Figure \ref{fig:FP}, while examples of misclassified matching pairs are reported in Figure \ref{fig:FN}.  From the examples reported it is possible to see that matching pairs on which our solution is very confident tend to be have the same pose and expression, suggesting that more powerful pose-normalization methods could further improve the verification accuracy. Some of the false negative examples are instead very difficult to classify since the subjects have had some sort of plastic surgery procedures (e.g. see Michael Jackson).

\begin{table*}
\caption{Average recognition accuracy comparison of our method and other baselines on the LAG dataset.}
\centering
\begin{tabular}{rllrr}
\toprule
ID & Method & Features & Accuracy & Improvement \\
\midrule
1.& Euclidean distance						& DCNN	     & 0,5917	&-0,1565 \\
2.& L2-norm \& Euclidean Distance	        & DCNN	 	 & 0,6498	&-0,0984 \\
3.& L2-norm \& Hellinger Distance			& DCNN		 & 0,6688	&-0,0794 \\
4.& Similarity Metric Learning (sub-SML) \cite{cao2013similarity}	& DCNN	 & 	0,7243	&-0,0239 \\
5.& OSS (one shot similarity) \cite{wolf2009one}& DCNN	        			 & 	0,6642	&-0,0840 \\
6.& Cosine Similarity	\cite{nguyen2011cosine} & DCNN	        			 & 	0,6508	&-0,0974\\
7.& Joint Bayesian \cite{chen2012bayesian} 		& DCNN	        			 & 	0,6633	&-0,0848\\
8.& CARC-NT	\cite{chen2014cross,chen2015face} 	& DCNN	        			 & 	0,7482	& -,----\\
9.& HDLBP \cite{chen2013blessing}	& LBP	                        		 & 	0,7153	& -0,0329 \\
\midrule
10.& Stacking (1 representation: DCNN)	& 1+2+3	 				& 0,6781	&-0,0701 \\
11.& Stacking (1 representation: DCNN)	& 1+2+3+4	 			& 0,7441	&-0,0041 \\
12.& Stacking (1 representation: DCNN)	& 1+2+3+4+5				& 0,7874&	0,0392 \\
13.& Stacking (1 representation: DCNN)	& 1+2+3+4+5+7			& 0,7712	&0,0231\\
14.& Stacking (1 representation: DCNN)	& 1+2+3+4+5+8			& 0,7921	&0,0439\\
\midrule
15.& Siamese DCNN								& DCNN	           & 0,7734	&0,0252 \\
16.& Stacking (1 representation: DCNN)	&1+2+3+4+5+8+15		   & 0,8010	&0,0528\\
17.& Siamese DCNN + feature injection					& DCNN with injected 1+2+3+4+5+8+15		           & 0,8277	&0,0795\\
18.& Stacking (2 representations: DCNN + LBP)	&1+2+3+4+5+8+9+15  & 0,8263	&0,0781\\
19.& \bf{Siamese DCNN + feaure injection}					& \bf{DCNN with injected	1+2+3+4+5+8+9+15}   & \bf{0,8495}	& \bf{0,1014}\\
\bottomrule
\end{tabular}
\label{tab:resultsLAG2}
\end{table*}

\begin{table}
\caption{Confusion matrix of our method on the LAG dataset.}
\centering
\begin{tabular}{lrr}
\toprule
			 & non-matching & matching \\
\midrule
non-matching & 0.8575  &  0.1425 \\
    matching & 0.1584  &  0.8416 \\
\bottomrule
\end{tabular}
\label{tab:errorDetails}
\end{table}

\section{Conclusions}
\label{sec:conclusion}
In this paper a new method for face verification across large age gaps is introduced along with a dataset containing variations of age in the wild. The proposed method exploits a deep convolutional neural network (DCNN) trained in a Siamese architecture ended with a contrastive loss function. The network discriminative power is further improved including a feature injection layer, which injects externally computed features into the deepest layers of the DCNN. 
Experimental results on the Large Age-Gap (LAG) dataset show that the proposed approach is able to outperform the face verification methods in the state of the art considered.

As future work we plan to extend this research in different directions:
\begin{itemize}
\item[-] Integration of the LAG dataset with the CAFE dataset \cite{liu2016deep} when it will be released, and further extension of the LAG dataset collecting more images for each identity.
\item[-] Investigation of the use of different face alignment algorithms such as \cite{hassner2015effective};
\item[-] Comparison and integration with face aging models \cite{geng2007automatic};
\item[-] Addition of pre-processing steps such as illuminant compensation \cite{bianco2014adaptive-color}
\item[-] Use of pre-classifiers to make verification task easier, e.g. age/gender/race classifier, expression classifier, etc.
\end{itemize}

\section*{Aknowledgments}
We gratefully acknowledge the support of NVIDIA Corporation with the donation of the Tesla K40 GPU used for this research.


\bibliographystyle{IEEEtran}
\bibliography{egbib}

\begin{thebibliography}{10}
\providecommand{\url}[1]{#1}
\csname url@samestyle\endcsname
\providecommand{\newblock}{\relax}
\providecommand{\bibinfo}[2]{#2}
\providecommand{\BIBentrySTDinterwordspacing}{\spaceskip=0pt\relax}
\providecommand{\BIBentryALTinterwordstretchfactor}{4}
\providecommand{\BIBentryALTinterwordspacing}{\spaceskip=\fontdimen2\font plus
\BIBentryALTinterwordstretchfactor\fontdimen3\font minus
  \fontdimen4\font\relax}
\providecommand{\BIBforeignlanguage}[2]{{%
\expandafter\ifx\csname l@#1\endcsname\relax
\typeout{** WARNING: IEEEtran.bst: No hyphenation pattern has been}%
\typeout{** loaded for the language `#1'. Using the pattern for}%
\typeout{** the default language instead.}%
\else
\language=\csname l@#1\endcsname
\fi
#2}}
\providecommand{\BIBdecl}{\relax}
\BIBdecl

\bibitem{jain2012face}
A.~K. Jain, B.~Klare, and U.~Park, ``Face matching and retrieval in forensics
  applications,'' \emph{IEEE multimedia}, vol.~19, no.~1, p.~20, 2012.

\bibitem{huang2007labeled}
G.~B. Huang, M.~Ramesh, T.~Berg, and E.~Learned-Miller, ``Labeled faces in the
  wild: A database for studying face recognition in unconstrained
  environments,'' Technical Report 07-49, University of Massachusetts, Amherst,
  Tech. Rep., 2007.

\bibitem{simonyan2014two}
K.~Simonyan and A.~Zisserman, ``Two-stream convolutional networks for action
  recognition in videos,'' in \emph{Advances in Neural Information Processing
  Systems}, 2014, pp. 568--576.

\bibitem{jiang2015exploiting}
Y.-G. Jiang, Z.~Wu, J.~Wang, X.~Xue, and S.-F. Chang, ``Exploiting feature and
  class relationships in video categorization with regularized deep neural
  networks,'' \emph{arXiv preprint arXiv:1502.07209}, 2015.

\bibitem{wang2015action}
L.~Wang, Y.~Qiao, and X.~Tang, ``Action recognition with trajectory-pooled
  deep-convolutional descriptors,'' \emph{arXiv preprint arXiv:1505.04868},
  2015.

\bibitem{zha2015exploiting}
S.~Zha, F.~Luisier, W.~Andrews, N.~Srivastava, and R.~Salakhutdinov,
  ``Exploiting image-trained cnn architectures for unconstrained video
  classification,'' \emph{arXiv preprint arXiv:1503.04144}, 2015.

\bibitem{perronnin2010improving}
F.~Perronnin, J.~S{\'a}nchez, and T.~Mensink, ``Improving the fisher kernel for
  large-scale image classification,'' in \emph{Computer Vision--ECCV
  2010}.\hskip 1em plus 0.5em minus 0.4em\relax Springer, 2010, pp. 143--156.

\bibitem{ng2015beyond}
J.~Y.-H. Ng, M.~Hausknecht, S.~Vijayanarasimhan, O.~Vinyals, R.~Monga, and
  G.~Toderici, ``Beyond short snippets: Deep networks for video
  classification,'' \emph{arXiv preprint arXiv:1503.08909}, 2015.

\bibitem{gers2003learning}
F.~A. Gers, N.~N. Schraudolph, and J.~Schmidhuber, ``Learning precise timing
  with lstm recurrent networks,'' \emph{The Journal of Machine Learning
  Research}, vol.~3, pp. 115--143, 2003.

\bibitem{parkcombining}
E.~Park, X.~Han, T.~L. Berg, and A.~C. Berg, ``Combining multiple sources of
  knowledge in deep cnns for action recognition,'' in \emph{Winter Conference
  on Applications of Computer Vision (WACV)}, 2016.

\bibitem{turk1991face}
M.~A. Turk and A.~P. Pentland, ``Face recognition using eigenfaces,'' in
  \emph{Computer Vision and Pattern Recognition, 1991. Proceedings CVPR'91.,
  IEEE Computer Society Conference on}.\hskip 1em plus 0.5em minus 0.4em\relax
  IEEE, 1991, pp. 586--591.

\bibitem{ahonen2006face}
T.~Ahonen, A.~Hadid, and M.~Pietikainen, ``Face description with local binary
  patterns: Application to face recognition,'' \emph{Pattern Analysis and
  Machine Intelligence, IEEE Transactions on}, vol.~28, no.~12, pp. 2037--2041,
  2006.

\bibitem{wright2009robust}
J.~Wright, A.~Y. Yang, A.~Ganesh, S.~S. Sastry, and Y.~Ma, ``Robust face
  recognition via sparse representation,'' \emph{Pattern Analysis and Machine
  Intelligence, IEEE Transactions on}, vol.~31, no.~2, pp. 210--227, 2009.

\bibitem{chen2013blessing}
D.~Chen, X.~Cao, F.~Wen, and J.~Sun, ``Blessing of dimensionality:
  High-dimensional feature and its efficient compression for face
  verification,'' in \emph{Computer Vision and Pattern Recognition (CVPR), 2013
  IEEE Conference on}.\hskip 1em plus 0.5em minus 0.4em\relax IEEE, 2013, pp.
  3025--3032.

\bibitem{taigman2014deepface}
Y.~Taigman, M.~Yang, M.~Ranzato, and L.~Wolf, ``Deepface: Closing the gap to
  human-level performance in face verification,'' in \emph{Proceedings of the
  IEEE Conference on Computer Vision and Pattern Recognition}, 2014, pp.
  1701--1708.

\bibitem{sun2014deep10k}
Y.~Sun, X.~Wang, and X.~Tang, ``Deep learning face representation from
  predicting 10,000 classes,'' in \emph{Proceedings of the IEEE Conference on
  Computer Vision and Pattern Recognition}, 2014, pp. 1891--1898.

\bibitem{sun2014deep}
Y.~Sun, Y.~Chen, X.~Wang, and X.~Tang, ``Deep learning face representation by
  joint identification-verification,'' in \emph{Advances in Neural Information
  Processing Systems}, 2014, pp. 1988--1996.

\bibitem{geng2007automatic}
X.~Geng, Z.-H. Zhou, and K.~Smith-Miles, ``Automatic age estimation based on
  facial aging patterns,'' \emph{Pattern Analysis and Machine Intelligence,
  IEEE Transactions on}, vol.~29, no.~12, pp. 2234--2240, 2007.

\bibitem{park2010age}
U.~Park, Y.~Tong, and A.~K. Jain, ``Age-invariant face recognition,''
  \emph{Pattern Analysis and Machine Intelligence, IEEE Transactions on},
  vol.~32, no.~5, pp. 947--954, 2010.

\bibitem{wu2012age}
T.~Wu and R.~Chellappa, ``Age invariant face verification with relative
  craniofacial growth model,'' in \emph{Computer Vision--ECCV 2012}.\hskip 1em
  plus 0.5em minus 0.4em\relax Springer, 2012, pp. 58--71.

\bibitem{li2011discriminative}
Z.~Li, U.~Park, and A.~K. Jain, ``A discriminative model for age invariant face
  recognition,'' \emph{Information Forensics and Security, IEEE Transactions
  on}, vol.~6, no.~3, pp. 1028--1037, 2011.

\bibitem{gong2013hidden}
D.~Gong, Z.~Li, D.~Lin, J.~Liu, and X.~Tang, ``Hidden factor analysis for age
  invariant face recognition,'' in \emph{Proceedings of the IEEE International
  Conference on Computer Vision}, 2013, pp. 2872--2879.

\bibitem{chen2014cross}
B.-C. Chen, C.-S. Chen, and W.~H. Hsu, ``Cross-age reference coding for
  age-invariant face recognition and retrieval,'' in \emph{Computer
  Vision--ECCV 2014}.\hskip 1em plus 0.5em minus 0.4em\relax Springer, 2014,
  pp. 768--783.

\bibitem{chen2015face}
B.-C. Chen, C.-S. Chen, and W.~Hsu, ``Face recognition and retrieval using
  cross-age reference coding with cross-age celebrity dataset,''
  \emph{Transactions on Multimedia}, vol.~17, no.~6, pp. 804--815, 2015.

\bibitem{liu2016deep}
L.~Liu, C.~Xiong, H.~Zhang, Z.~Niu, M.~Wang, and S.~Yan, ``Deep aging face
  verification with large gaps,'' \emph{Multimedia, IEEE Transactions on},
  vol.~18, no.~1, pp. 64--75, 2016.

\bibitem{kumar2009attribute}
N.~Kumar, A.~C. Berg, P.~N. Belhumeur, and S.~K. Nayar, ``Attribute and simile
  classifiers for face verification,'' in \emph{Computer Vision, 2009 IEEE 12th
  International Conference on}.\hskip 1em plus 0.5em minus 0.4em\relax IEEE,
  2009, pp. 365--372.

\bibitem{yi2014learning}
D.~Yi, Z.~Lei, S.~Liao, and S.~Z. Li, ``Learning face representation from
  scratch,'' \emph{arXiv preprint arXiv:1411.7923}, 2014.

\bibitem{cootes2008fg}
T.~Cootes and A.~Lanitis, ``The fg-net aging database,'' 2008.

\bibitem{ricanek2006morph}
K.~Ricanek~Jr and T.~Tesafaye, ``Morph: A longitudinal image database of normal
  adult age-progression,'' in \emph{Automatic Face and Gesture Recognition,
  2006. FGR 2006. 7th International Conference on}.\hskip 1em plus 0.5em minus
  0.4em\relax IEEE, 2006, pp. 341--345.

\bibitem{viola2004robust}
P.~Viola and M.~J. Jones, ``Robust real-time face detection,''
  \emph{International journal of computer vision}, vol.~57, no.~2, pp.
  137--154, 2004.

\bibitem{kazemi2014one}
V.~Kazemi and J.~Sullivan, ``One millisecond face alignment with an ensemble of
  regression trees,'' in \emph{Computer Vision and Pattern Recognition (CVPR),
  2014 IEEE Conference on}.\hskip 1em plus 0.5em minus 0.4em\relax IEEE, 2014,
  pp. 1867--1874.

\bibitem{bromley1993signature}
J.~Bromley, J.~W. Bentz, L.~Bottou, I.~Guyon, Y.~LeCun, C.~Moore,
  E.~S{\"a}ckinger, and R.~Shah, ``Signature verification using a “siamese”
  time delay neural network,'' \emph{International Journal of Pattern
  Recognition and Artificial Intelligence}, vol.~7, no.~04, pp. 669--688, 1993.

\bibitem{chopra2005learning}
S.~Chopra, R.~Hadsell, and Y.~LeCun, ``Learning a similarity metric
  discriminatively, with application to face verification,'' in \emph{Computer
  Vision and Pattern Recognition, 2005. CVPR 2005. IEEE Computer Society
  Conference on}, vol.~1.\hskip 1em plus 0.5em minus 0.4em\relax IEEE, 2005,
  pp. 539--546.

\bibitem{cao2013similarity}
Q.~Cao, Y.~Ying, and P.~Li, ``Similarity metric learning for face
  recognition,'' in \emph{Computer Vision (ICCV), 2013 IEEE International
  Conference on}.\hskip 1em plus 0.5em minus 0.4em\relax IEEE, 2013, pp.
  2408--2415.

\bibitem{wolf2009one}
L.~Wolf, T.~Hassner, and Y.~Taigman, ``The one-shot similarity kernel,'' in
  \emph{Computer Vision, 2009 IEEE 12th International Conference on}.\hskip 1em
  plus 0.5em minus 0.4em\relax IEEE, 2009, pp. 897--902.

\bibitem{nguyen2011cosine}
H.~V. Nguyen and L.~Bai, ``Cosine similarity metric learning for face
  verification,'' in \emph{Computer Vision--ACCV 2010}.\hskip 1em plus 0.5em
  minus 0.4em\relax Springer, 2011, pp. 709--720.

\bibitem{chen2012bayesian}
D.~Chen, X.~Cao, L.~Wang, F.~Wen, and J.~Sun, ``Bayesian face revisited: A
  joint formulation,'' in \emph{Computer Vision--ECCV 2012}.\hskip 1em plus
  0.5em minus 0.4em\relax Springer, 2012, pp. 566--579.

\bibitem{wolpert1992stacked}
D.~H. Wolpert, ``Stacked generalization,'' \emph{Neural networks}, vol.~5,
  no.~2, pp. 241--259, 1992.

\bibitem{hassner2015effective}
T.~Hassner, S.~Harel, E.~Paz, and R.~Enbar, ``Effective face frontalization in
  unconstrained images,'' in \emph{Computer Vision and Pattern Recognition
  (CVPR), 2015 IEEE Conference on}.\hskip 1em plus 0.5em minus 0.4em\relax
  IEEE, 2015, pp. 4295--4304.

\bibitem{bianco2014adaptive-color}
S.~Bianco and R.~Schettini, ``Adaptive color constancy using faces,''
  \emph{IEEE Transactions on Pattern Analysis and Machine Intelligence},
  vol.~36, no.~8, pp. 1505--1518, 2014.

\end{thebibliography}
%
%
%

%


\begin{IEEEbiographynophoto}{Simone Bianco}
received the BSc and the MSc degrees in mathematics from the University of Milano-Bicocca, Italy, in 2003 and 2006, respectively. He received the PhD degree in computer science from the Department of Informatics, Systems and Communication of the University of Milano-Bicocca, Italy, in 2010, where he is currently a post-doc researcher. His research interests include computer vision, machine learning, optimization algorithms, and color imaging.
\end{IEEEbiographynophoto}

%
%




\end{document}